\documentclass[11pt]{article}
\usepackage[a4paper]{geometry}
\usepackage{clic2017}
\usepackage{times}
\usepackage{url,multicol}
\usepackage{latexsym}
\usepackage{soul}

\usepackage{xcolor}

\newcommand{\seqkd}[0]{\texttt{Seq-KD}}
\newcommand{\seqinter}[0]{\texttt{Seq-Inter}}
\newcommand{\wordkd}[0]{\texttt{Word-KD}}

\newcommand\blfootnote[1]{%
  \begingroup
  \renewcommand\thefootnote{}\footnote{#1}%
  \addtocounter{footnote}{-1}%
  \endgroup
}

\title{On Knowledge Distillation for Direct Speech Translation}

\author{Marco Gaido\textsuperscript{1,2},
Mattia Antonino Di Gangi\textsuperscript{3}\textsuperscript{*},  
Matteo Negri\textsuperscript{1}, Marco Turchi\textsuperscript{1} \\
  \textsuperscript{1}Fondazione Bruno Kessler, Trento, Italy \\
  \textsuperscript{2}University of Trento, Trento, Italy \\
  \textsuperscript{3}AppTek, Aachen, Germany \\
  {\tt \{mgaido,digangi,negri,turchi\}@fbk.eu}}

\date{}

\begin{document}
\maketitle
\begin{abstract}
  \textbf{English.}  Direct speech translation (ST)
  has shown to be a complex task requiring knowledge
  transfer from its sub-tasks: automatic speech
  recognition (ASR) and machine translation (MT).
  For MT, one of the most promising techniques
  to transfer knowledge is knowledge distillation.
  In this paper, we compare the different solutions
  to distill knowledge in a sequence-to-sequence
  task like ST.
  Moreover, we analyze eventual drawbacks of this approach and
  how to alleviate them maintaining the benefits in terms of translation quality.
\end{abstract}

\begin{abstract-alt}
 \textrm{\bf{Italiano.}} \`E stato dimostrato che
 la \textit{speech translation} (ST) diretta è un'operazione
 complessa che richiede l'adozione di tecniche di
 \textit{knowledge transfer} sia da \textit{automatic speech recognition} (ASR)
 che da \textit{machine translation} (MT).
 Per quanto riguarda MT, una delle
 tecniche più promettenti è la \textit{knowledge distillation} (KD). In questo lavoro, confrontiamo
 diverse possibili soluzioni di KD per addestrare modelli sequence-to-sequence come quelli di ST. Inoltre,
 analizziamo eventuali problemi causati
 da questa tecnica e come attenuarli mantenendo i benefici in termini di qualità della traduzione.
\end{abstract-alt}

\section{Introduction}

\blfootnote{\textsuperscript{*}Work done during the PhD at Fondazione Bruno Kessler and University of Trento.}
\blfootnote{Copyright \copyright 2020 for this paper by its authors. Use permitted under Creative Commons License Attribution 4.0 International (CC BY 4.0).}
%Direct speech translation (ST) 
Speech translation (ST) 
refers to the process of translating utterances in one language into text in a different language.
Direct ST is an emerging paradigm that consists in translating
without intermediate representations \cite{berard_2016,weiss2017sequence}.
It is a newer and alternative approach to cascade solutions \cite{StentifordSteer88,Waibel1991b},
in which the input audio is 
first transcribed with an automatic speech recognition (ASR) model and then the transcript is translated into
the target language with a machine translation (MT) model.

The rise of 
%direct ST models
the direct ST paradigm
is motivated by 
its theoretical
and practical advantages, namely:
\textit{i)} during the translation phase it has access
to information present in the audio that is lost in its transcripts (eg. prosody,
characteristic of the speaker\footnote{For instance, the pitch of the voice is
a cue for the sex of the speaker. Although the gender is a social aspect and
does not depend on physical attributes, in many cases sex and gender coincide,
so systems relying on this are likely to have a better accuracy than those
that do not have access to any information regarding the speaker \cite{bentivogli-etal-2020-gender,gaido-etal-2020-gender}.}),
\textit{ii)} there is no \textit{error propagation} (in cascade systems the
errors introduced by the ASR are propagated to the MT, which has no cues to recover them),
\textit{iii)} the latency is lower (as data flows through a single system instead of two), and
\textit{iv)} the management is easier (as there is a single model to maintain and no integration between 
separate
modules is needed).

On the downside,
direct ST suffers from the lack of large ST training corpora.
This problem has been addressed by researchers through transfer learning from the high-resource sub-tasks \cite{berard2018end,bansal-etal-2019-pre,liu2019endtoend},
multi-task trainings \cite{weiss2017sequence,anastasopoulos-2018-multitask,bahar2019comparative},
and the proposal of data augmentation techniques \cite{jia2018leveraging,bahar-2019-specaugment,nguyen2019improving}.
%Training direct ST systems proved to require knowledge transfer from ASR and MT to achieve good results \cite{bahar2019comparative}.
%The knowledge is usually transferred through component initialization, but for MT another possibility is knowledge distillation.
In this work, we focus on the transfer learning from MT.
The classic approach consists in pre-training the decoder with that of an MT model.
Its benefit,
however, is controversial: indeed, \cite{bahar2019comparative} showed that it is effective only with the addition of an \textit{adapter}
layer, but this has not been confirmed in \cite{gaido-etal-2020-end}, while in \cite{inaguma-etal-2020-espnet} it always brought improvements. Another, more promising possibility consists in distilling knowledge from an MT model.

Knowledge distillation (KD)
is a knowledge transfer technique introduced for
model compression \cite{hinton2015distilling}. A 
small \textit{student} model
is trained computing the KL-divergence~\cite{kullback1951} with the output
probability distribution of a big
\textit{teacher} model.
Although KD was introduced in the context of image processing,
its effectiveness suggested its adoption in other fields.
Specifically,
\newcite{liu2019endtoend}
showed that
using an MT system as teacher
brings significant improvements
to direct ST models.
However, they did not compare the different methods
to distill knowledge in
sequence-to-sequence models \cite{kim2016sequencelevel}
and they did not analyze
possible negative
effects of 
adopting this technique.

In this paper, we analyze 
different sequence-to-sequence KD techniques (\textit{word level}, \textit{sequence-level}, \textit{sequence interpolation}) and their 
combination in the context of direct ST.
Then, we study the effect of the best technique
on a strong system trained on a large amount of data
to reach state-of-the-art results.
We show that word-level KD is the best approach and that fine-tuning
the resulting model without KD brings further improvements.
Finally, we analyze the limitations and the problems
present in models trained with KD, which
are partly solved by the final finetuning.

\section{Sequence-level Knowledge Distillation}

We focus on distilling knowledge from an MT model
to an ST model.
This is helpful due to the better
results achieved by MT, which is an easier task than ST, as it does not involve the recognition of the audio content, and it also benefits from the availability of large training corpora.
Our student (ST) model is trained to produce the same output distribution of the teacher (MT) model when the latter is fed with the transcript
of the utterances passed as input to the ST model. 
As KD was introduced in the context of 
\textit{classification}
tasks, 
while
ST and MT are sequence-to-sequence \textit{generation} tasks,
an adaptation is required 
for its application.
\newcite{kim2016sequencelevel} introduced three methods to distill knowledge in sequence-to-sequence models:
\textit{i)} word-level KD,
\textit{ii)} sequence-level KD,
and \textit{iii)} sequence interpolation.

\noindent\textbf{Word-level KD} (\wordkd) refers to computing the KL-divergence between the distribution of the teacher and student models on each token to be predicted.
As recomputing the teacher output 
at each iteration is computationally
expensive (it needs a forward pass of the MT model),
we explored the possibility to
pre-compute and store the teacher outputs. To this aim, we experimented with truncating the output distribution to have a lower memory footprint,
as proposed in MT \cite{tan2018multilingual}.

\noindent\textbf{Sequence-level KD} (\seqkd) consists in considering as target the output generated by the teacher model using the beam search. 

\noindent\textbf{Sequence interpolation} (\seqinter) is similar to Seq-KD, but
the target is the sentence with the highest BLEU score \cite{papineni-2002-bleu} with respect to the ground truth among the n-best generated by the beam search with the teacher model.

%We
As done in \cite{kim2016sequencelevel}, we
also combine these methods to analyze whether they are complementary or not. Finally, we experiment with fine-tuning the model trained with KD on
the reference translations.

\section{Experimental Settings}

We performed preliminary experiments on a limited amount of data to compare
the three KD methods. Then, we created a model
exploiting all the available corpora
with the best technique
to analyze the KD behavior
in a real scenario.

\subsection{Data}

We first experiment using only Librispeech \cite{librispeech}, an ST corpus with English audio, transcripts and French translations.
We
use the \textit{(audio, transcript)} pairs for the ASR pre-training, the \textit{(transcript, translation)} pairs 
to train
the MT teacher,
and the \textit{(audio, translation)} pairs for the ST training.

Then, we 
built
an English-Italian model. In addition to Librispeech, the ASR pre-training involves 
TED-LIUM 3 \cite{Hernandez_2018},
Mozilla Common Voice,\footnote{\url{https://voice.mozilla.org/}}
How2 \cite{sanabria18how2} and the en-it section of MuST-C \cite{mustc}. The MT teacher is trained on the OPUS datasets \cite{opus}, cleaned using the ModernMT framework~\cite{modernmt}.\footnote{With the \texttt{CleaningPipelineMain} class.}
For ST, we use
the en-it section of MuST-C and Europarl-ST \cite{europarlst}.

%We extract 40 Mel filter banks from the audio using a 25 ms window and 10 ms slide.
%Speaker normalization is 
%performed using XNMT \cite{neubig-etal-2018-xnmt}.
We pre-process the input audio extracting a 40-dimensional feature vector from a span of 25 ms every 10 ms using Mel filter bank. During this pre-processing performed with XNMT \cite{neubig-etal-2018-xnmt}, we also apply speaker normalization.
The text is tokenized and the punctuation is normalized with Moses~\cite{koehn-etal-2007-moses}.
We create 8,000 shared BPE merge rules on the MT data of each experiment and apply them to divide the text into sub-word units.
Samples lasting more than 20 seconds are discarded in order to avoid out of memory issues during 
training.

\subsection{Models}

For ST and ASR we use the S-Transformer architecture \cite{di-gangi-etal-2019-enhancing,digangi2019adapting} with logarithmic distance penalty in the encoder.
In particular, in the experiments on Librispeech we train a small model using the basic configuration by \newcite{di-gangi-etal-2019-enhancing}, while in the experiment with all the data we follow the BIG configuration.
In the second case,
we also slightly modify the architecture to improve performance by removing the 2D attention layers
%Moreover,
%we change
and changing
 the number of Transformer Encoder layers and Transformer Decoder layers to be respectively 11 and 4 in ST and 8 and 6 in the ASR pre-training \cite{gaido-etal-2020-end}.
The different number of layers between ASR and ST
is motivated by the idea of having adaptation layers \cite{jia2018leveraging,bahar2019comparative}.

For MT we use a
Transformer with 6 layers for
both the encoder and the decoder.
In the preliminary experiments, we use a
small model
with 512 hidden features in the attention layers,
2,048 hidden units in the feed-forward layers
and 8 attention heads; in the experiment with more data we double all these parameters.

\subsection{Training}

We optimize our models with Adam \cite{adam} using betas (0.9, 0.98).
The learning rate increases linearly for 4,000 steps starting from 1e-7 to 5e-3. Then it decays according to the inverse square root policy.
In fine-tunings, the learning rate is fixed at 1e-4.
A 0.1 dropout is applied and the total batch size is 64.
When we do not use KD, the loss is label smoothed cross entropy \cite{szegedy2016rethinking} with 0.1 smoothing factor.

In the final training with 
all the data,
we
apply SpecAugment \cite{Park_2019} with probability 0.5, 13 \textit{frequency masking pars},
20 \textit{time masking pars}, 2 \textit{frequency masking num}, and 2 \textit{time masking num}. 
We also
increase the overall batch size to 512.
Moreover, the ASR pre-training is performed as a multi-task training in which we add a CTC loss (predicting the output transcripts) on the encoder output \cite{kim2016joint}.

Our code is based on the Fairseq library \cite{ott-etal-2019-fairseq}, which relies on PyTorch \cite{paszke2017automatic}, and it is available open source at \url{https://github.com/mgaido91/FBK-fairseq-ST}. The models are trained on 8 GPU K80 with 11 GB of RAM.

\section{Results}

First, we experiment truncating the output distribution generated by the teacher model.
Table \ref{tab:libri_topk} shows that truncating
the output to few top tokens does not affect
significantly the performance. On the contrary,
the best result is obtained using the top 8 tokens.
Hence, all our experiments with \wordkd{} use the top 8 tokens of the teacher.

\begin{table}[ht]
\small
\centering
\begin{tabular}{r|r}
Top K & BLEU  \\
\hline
4 & 16.43 \\
8 & \textbf{16.50} \\
64 & 16.37 \\
1024 & 16.34 \\
\end{tabular}
\caption{Results %on Librispeech
with different $K$ values, where $K$ is the number of tokens considered for \wordkd.}
\label{tab:libri_topk}
\end{table}

Then, we try different values for the temperature $T$ parameter.
The temperature is a parameter 
used to
sharpen
(if $T < 1$) or soften (if $T > 1$) the output distribution.
In particular, by adding the temperature, the $softmax$ function that converts the \textit{logits} $z_i$ into probabilities $p_i$ becomes:

\begin{equation}
  p_i = \frac{e^{z_i / T}}{\sum (e^{z_i / T})}
  \label{gating_lambgda_eq}
\end{equation}

%
%Using a
A higher temperature has been claimed to help learning the so-called \textit{dark knowledge} \cite{hinton2015distilling},
one of the possible reasons alluded to justify the success of KD.
Indeed, with a high temperature, the cost function is similar to minimizing the squared distance between the logits produced by the student and teacher networks.
So logits with very negative values -- which are basically ignored with low temperature -- become important to be learnt by the student network.
For a demonstration, please refer to \cite{hinton2015distilling}.
Table \ref{tab:libri_temp}
reports the BLEU score for different values of $T$ and
indicates that 
the default $T = 1$ is the best value.
This result 
%suggest
suggests
that, in ST, the networks do not have the capacity of MT models trained on the same data. So focusing on the mode of the probability distribution works best.

\begin{table}[ht]
\small
\centering
\begin{tabular}{r|r}
$T$ & BLEU  \\
\hline
1.0 & \textbf{16.50} \\
4.0 & 16.11 \\
8.0 & 14.27 \\
\end{tabular}
\caption{Results 
with different temperatures ($T$).}
\label{tab:libri_temp}
\end{table}

\begin{table}[ht]
\small
\centering
\begin{tabular}{l|r}
 & BLEU  \\
\hline
Baseline & 9.4 \\
\hline
\wordkd & 16.5 \\
\seqkd & 13.4 \\
\seqinter & 13.3 \\
\hline
\seqkd{} + \wordkd{} & 15.7 \\
\hline
\wordkd{} + FT \seqkd{} & 16.7 \\
\seqkd{} + FT \wordkd{} & \textbf{16.8} \\
\wordkd{} + FT w/o KD & \textbf{16.8} \\
\end{tabular}
\caption{Results
of the small model on Librispeech 
with different KD methods and 
%\mg{combinations}.
combining them in a single training or in consecutive trainings through 
a fine-tuning (FT).
}
\label{tab:libri_res}
\end{table}

\begin{table*}[htp]
\centering
\small
\begin{tabular}{l|c|ccc|ccc|c}
\hline
% \multicolumn{1}{c|}{} & \multicolumn{4}{c|}{\textbf{En-it}} \\ \cline{2-5}
  & & \multicolumn{3}{c}{\textbf{Female}} & \multicolumn{3}{c|}{\textbf{Male}} & \textbf{Bias}  \\
  \hline
  & BLEU & Corr. & Wrong & Diff. & Corr. & Wrong & Diff. & Diff. M - Diff. F \\
\hline
Base ST \cite{bentivogli-etal-2020-gender}  &  21.5 &	\textbf{26.7} & 27.2 & \textbf{-0.5} & \textit{46.3} & 6.8 & \textit{39.5} & \textbf{40.0} \\
\hline
MT  & \textbf{30.3} &  10.8 &	55.5 &	-44.7 &	\textbf{54.4} &	7.1 &	\textbf{47.3} & 92.0 \\
\seqkd{} + \wordkd{} + FT \wordkd{}  & 22.8 & 12.3 &	46.5 &	-34.2 &	45.4 &	8.1 &	37.3 & 71.5 \\
\hspace{8mm} + FT w/o KD & \textit{27.7} & 19.8 &	39.0 &	-19.2 &	43.2 &	10.5 &	32.7 & 51.9 \\
\hline
\end{tabular}
\caption{Accuracy on Category 1 of the MuST-SHE test set of a base direct ST model and models created using KD.
A high \textit{Diff.} means that the model is able to recognize the speaker's gender and the gap between the \textit{Diff.} on the two genders indicates the bias towards one of them. The reported BLEU score refers to the MuST-C test set and shows the translation quality of the model.}
  \label{tab:accuracy-cat1}
\end{table*}

Then, we compare the different sequence-level KD techniques. 
We also combine them either in the same training or
in consecutive trainings through a fine-tuning (FT).
The results are presented in Table \ref{tab:libri_res}.
We can notice that all the methods improve significantly over the baseline:
KD makes the training easier and more effective. Among them, \wordkd{} achieves the best results by a large margin.
Combining it with another method in the same training is harmful (\seqkd{} + \wordkd{}), while a fine-tuning on a different KD method or without KD (i.e. using the ground-truth target and label smoothed cross entropy) improves results by up to 0.3 BLEU (\seqkd{} + FT \wordkd{} and \wordkd{} + FT w/o KD).
These results confirm the choice by \cite{liu2019endtoend}, but differ from those of \cite{kim2016sequencelevel}. So, we can conclude that the best sequence-to-sequence KD technique is task-dependent and that the best option to distill knowledge from MT to ST is the word-level KD.

To validate the effectiveness of KD in a real case, we create a model translating English utterances into Italian text leveraging all the available corpora for each task. Our ASR pre-trained model scores 10.21 WER on the MuST-C test set, while the teacher MT model scores 30.3 BLEU on the Italian reference for same test set.
We train our ST model first on the ASR corpora for which we generated the target with the MT model (resulting in a \seqkd{} + \wordkd{} training).
Note
that we could not use this data without \seqkd{} or \seqinter{}, hence we opted for the best training including 
one of them (\seqkd{} + \wordkd{}). Second, we fine-tune the model on the ST corpora with \wordkd{}. Third, we fine-tune without KD as in the
case leading to the best result
(Table \ref{tab:libri_res}).
So, our training is:
\seqkd{} + \wordkd{} (on ASR data) + FT \wordkd{} + FT w/o KD.
After the first two steps, our ST model scores 22.8 BLEU on the MuST-C test set, while after the final fine-tuning 
the result is
it scores 27.7 BLEU.
This highlights the importance of fine-tuning without KD.

\section{Analysis}

We analyze the outputs of the en-it model to assess whether,
despite the benefits in terms of translation quality, KD introduces limitations or issues.
Namely, we checked whether the lack of access of the MT teacher
to information present in the audio and not in the text
(such as the gender\footnote{This is true if the gender identity coincides with the biological sex. This assumption holds true in nearly all our data.} of the speaker)
hinders the ability of the final model to exploit such
knowledge.
Moreover, we compared the output generated by the model before 
fine-tuning without KD and after it
to determine the reasons of the significant BLEU improvement.

Direct ST systems have been shown to be able to exploit the audio
to determine the gender of the speaker and reflect it better
in the translations into languages rich of gender marked words
\cite{bentivogli-etal-2020-gender}.
This is not possible for an MT system
that has no clue
regarding the speaker's gender.
We tested the performance of our models on the category 1
of the
MuST-SHE test set \cite{bentivogli-etal-2020-gender}
(which contains gender marked word related to the speaker)
to check whether distilling knowledge from MT harms this advantage of ST systems or not.
Table \ref{tab:accuracy-cat1} shows that, indeed, systems trained with KD inherit the bias
from the MT system and, although the final fine-tuning mitigates the issue,
the final model has a higher gender bias than a base ST system without KD (regarding the words related to the speaker).

The
better translation of speaker's gender marked words does not explain
the big BLEU improvement obtained with 
fine-tuning. Hence, we performed a manual analysis
of sentences with the highest TER \cite{snover-ter-2006} reduction.
The analysis revealed three main
types of 
enhancements, with the first being the most significant.

\noindent \textbf{Samples with multiple sentences.}
Some utterances contain more than one sentence. In this case, the model trained with KD
tends to generate the translation of only the first sentence,
%and truncate after it,
ignoring the others.
This is likely caused by
the fact that MT training data is mostly sentence-level. For this reason, the MT model tends to assign a
%learning a too
high probability of the \texttt{EOS} symbol after the dot. The student ST model learns to mimic this harmful behavior and, as in ST training and test samples often include more than one sentence, to wrongly truncate the generation once the first sentence is completed.
%from the MT model.
The fine-tuned model, instead,
generates all the sentences.

\noindent \textbf{Verbal tenses.} The fine-tuned model tends to produce the
correct verbal tense,
while before the fine-tuning the verbal tense is often not precise,
likely because the MT model favors more generic forms.
For instance,
\textit{``That meant I was going to be on television''} should be translated as
\textit{``Significava che \textbf{sarei andata} in televisione''}.
%, but the
The model before fine-tuning produces
\textit{``Questo significava che \textbf{stavo andando} in tv''} while the fine-tuned model uses the correct verbal tense  \textit{``Questo significava che \textbf{sarei andata} in televisione''}.
Despite relevant for the final score, it is debatable whether this is a real improvement of the fine-tuned model, as in some cases both verbal tenses are acceptable or their correctness depends on the context (e.g. in informal conversations, the usage of conjunctive forms is often replaced with indicative tenses).

\noindent \textbf{Lexical choices.} In some cases, the fine-tuned model chooses more appropriate words, probably thanks to the fine-tuning on in-domain data.
For instance, the reference translation for \textit{``She has taken a \textbf{course} in a business school, and she has become a veterinary doctor''}
is \textit{``Ha seguito un \textbf{corso} in una scuola di business, ed è diventata una veterinaria''}.
The corresponding utterance
was translated by the model before the fine-tuning into \textit{``Ha frequentato una \textbf{lezione} di economia ed è diventata una dottoressa veterinaria''}, while after the fine-tuning the translation is \textit{``Ha frequentato un \textbf{corso} in una business school, ed è diventata una dottoressa veterinaria''}.

%\mt{Secondo me manca una frase che riassume un po' tutto.}
We can conclude that KD provides a benefit in terms of overall translation quality, but the resulting ST system also learns negative behaviors (such as the masculine default for the speaker-related words that exacerbates the gender bias). These are partly solved by performing a fine-tuning without KD, which keeps (and even enhances) on the other side the translation capabilities.

\section{Conclusions}

%In this work we presented the huge benefits in terms of translation quality
%In this work, we
We presented and analyzed the benefits and issues
brought by distilling knowledge from an MT system for direct ST models.
We compared the different KD techniques and our experiments indicated that
the best training procedure consists in a pre-training with word-level KD and a fine-tuning
without KD.
%Finally,
Then,
we showed that KD from MT models
%can lead to
causes an increased gender bias, omission of sentences in multi-sentential utterances and more generic word/verbal-tense choices. Finally, we demonstrated
%and
that a fine-tuning helps resolving these issues, although the exacerbation of gender bias is not solved, but only alleviated.

\section*{Acknowledgments}
This work is part of the ``End-to-end Spoken Language Translation in Rich Data Conditions'' project,\footnote{\url{https://ict.fbk.eu/units-hlt-mt-e2eslt/}} which is financially supported by an Amazon AWS ML Grant.

\bibliographystyle{acl}
\bibliography{clic2020}

\end{document}